# A Semantic approach for effective document clustering using WordNet

Mrs.Leena H. Patil, Dr. Mohammed Atique

*Abstract*— Now a days, the text document is spontaneously increasing over the internet, e-mail and web pages and they are stored in the electronic database format. To arrange and browse the document it becomes difficult. To overcome such problem the document preprocessing, term selection, attribute reduction and maintaining the relationship between the important terms using background knowledge, WordNet, becomes an important parameters in data mining. In these paper the different stages are formed, firstly the document preprocessing is done by removing stop words, stemming is performed using porter stemmer algorithm, word net thesaurus is applied for maintaining relationship between the important terms, global unique words, and frequent word sets get generated, Secondly, data matrix is formed, and thirdly terms are extracted from the documents by using term selection approaches tf-idf, tf-df, and tf2 based on their minimum threshold value. Further each and every document terms gets preprocessed, where the frequency of each term within the document is counted for representation. The purpose of this approach is to reduce the attributes and find the effective term selection method using WordNet for better clustering accuracy. Experiments are evaluated on Reuters Transcription Subsets, wheat, trade, money grain, and ship, Reuters 21578, Classic 30, 20 News group (atheism), 20 News group (Hardware), 20 News group (Computer Graphics) etc.

*Index Terms*— Introduction, Document Preprocessing, WordNet, Term Selection approach, Experimental results.

## I. INTRODUCTION

A substantial portion of the available information is stored in Text databases, which consist of large collections of documents from various sources, such as news articles, research papers, books, digital libraries, e-mail messages and web pages. [1]Text documents are growing rapidly due to the increasing amount of information available in electronic and digitized form, such as electronic publications, various kinds of electronic documents, e-mail, and the World Wide Web. Nowadays most of the information regarding government, industry, business, and other institutions are stored electronically, in the form of text databases. Most of the text databases are semi structured data format which they are neither completely structured. Document Preprocessing and clustering is very useful tool in today's world where large amount of documents and information are stored and retrieved electronically. As text data are inherently unstructured, some researchers applied different technique for document management. Researchers has presented knowledge discovery in text system, which uses the simplest information extraction to get interesting information and knowledge from unstructured text collection. [10]To enable the representation process and effective transformation, the word frequencies require to be normalized in terms of their relative frequencies which are present in document and over the entire collection. To organize the large document corpus making later navigating, the document browsing becomes more easy, friendly and efficient. Almost, it is impossible for the human being to read through all the text documents and find out the relative for a specific topic and how to organize large document. To organize the large amount of data and stored in a structured formats certain data mining techniques are able to use or extract the necessary information from the unstructured document collections. [9][10]Because of this, text mining techniques are useful in processing these documents. The goal of text mining is to structure document collections to improve the ability of users to retrieve and apply knowledge implicitly contained within those collections. Text mining proceeds through different phases to complete the goal: pre-processing, applying WordNet and term selection approach. Attributes and dimension reduction are the important parameters in text mining. This improves the performance of term selection methods by reducing dimensions so that text mining procedures process data with a reduced number of terms and accuracy get improved.

WordNet is the product of a research project in Princeton University which has attempted to model the lexical knowledge of a native speaker of English [12]. In WordNet Nouns, verbs, adjectives, and adverbs are connected to each other into hierarchies by well-defined types of semantic relations. These semantic relations for nouns include Hyponym/Hypernym (is-a), Part Meronym/Part Holonym (part-of), Member Meronym/Member Holonym (member-of), Substance Meronym/Substance Holonym (substance-of) and so on. For example, a car is a wheeled vehicle (is-a), and a cell is part of organism (part-of). Hyponym/hypernym (is-a) is the most common relations. [13]Language semantics are mostly captured by nouns and noun phrases, in this the similarity measures based on nouns and is-a relations in WordNet is discussed. To overcome such problem we design an algorithm which will deal and work on large scale data sets with high dimensions.

Prof. Mrs. Leena H. Patil, Computer Science and Engineering, Amravati University/ SGBAU / SGBAU CSE Deptt, Amravati, India, 9373612549, (e-mail: harshleena23@rediffmail.com).

Dr. Mohamed Atique, Computer Science and Engineering, Amravati University/ SGBAU/ SGBAU CSE Deptt., Amravati, India ,09011039313, (e-mail: atique_shaikh1@rediffmail.com).

In this approach document are preprocessed, dimension get reduced, WordNet is applied for arranging words as noun, verb, adverb, and adjectives, for further having better clustering, term selection methods are used. Documents get preprocessed by several steps: Firstly stopwords are removed, Secondly stemming is perform by using porter's stemmer algorithm, Thirdly WordNet senses is applied, Global unique words and frequent word set gets generated by using feature selection approaches tf-idf, tf-df, tf2. Finally the attribute reduction methods are used based on their entropy and significant measures by using forward greedy attribute reduction algorithm. Experimental evaluations are performed on the above approaches on Reuters transcription subset, wheat, grain, ship, money, trade datasets, Reuters 21578, Classic 30, 20 News group (atheism), 20 News group (Hardware), 20 News group (Computer Graphics) etc. From these evaluations the best and better approach for document clustering is establish

## II. DOCUMENT PREPROCESSING

To Organize and browse thousands of documents smoothly, document preprocessing becomes an most important step, which affects on the result. It describes the required transformation processing of documents to obtain the designated representation of documents. [7]Thousands of words are present in a document set, the aim of this is to reduce dimensionality for having the better accuracy for classification.
Document preprocessing are divided into following stages:
1. Each sentences gets divided into terms
2. Stop words are removed: Stopword List is used that contains the words to be excluded. The Stopword list is applied to remove terms that have a special meaning but do not discriminate for topics
3. Word Stemming: Developed stemming algorithm such as porters is used to reduce a word to its stem or root form.
4. WordNet Senses Disambiguation is applied as an English Database.
5. Global Unique words and frequent word set gets generated.

### A. Stop-Word Removing

Stop-words are words that from non-linguistic view do not carry information. Stop-words remove the non-information behavior words from the text documents and reduce noisy data. [9]One major property is that there are extremely common words present and the explanation of the sentences still held after these stop-words are removed. Most existent search engines do not record stop-words in order to reduce the space and speed up the searches. To organize large corpus, removing the stop words affords the similar advantages. Firstly it could save huge amount of space. Secondly it helps to deduce the noises and keep the core words, and it will make later processing more effective and efficient.

### B. Stemming

This process is used for transforming the words into their stem. In many languages the various syntactic form of words are used and explain with the same basic concepts. The most important technique called stemming is used for the reduction of words into their root. Many words from the English language can be reduced to their base form or stem e.g. agreed, agreeing, disagree, agreement and disagreement belong to agree. Porter Stemmer is a widely applied method to stem documents. [11]It is compact, simple and relatively accurate. It does not require creating a suffix list before applied.

### C. Wordnet

Using Lexical database the WordNet approach measures the relatedness of terms from the words. Based on these, we can compute scores of semantic relatedness of terms found from WordNet. In general as a dictionary, WordNet covers some specific terms from every subject related to their terms. Wordnet as a lexical database map all the stemmed words from the standard documents into their specifies lexical categories. However, in different fields, the semantic relation of terms may be different as proposed in [13]. In this approach the WordNet 2.1 is used which contains 41 lexical categories as nouns and verbs. For example, the word "dog" and "cat" both belong to the same category "noun.animal". Some words also has multiple categories like word "Washington" has 3 lexical categories (noun.location, noun.group, noun.person) because it can be the name of the American president, the city place, or a group in the concept of capital.

## III. FEATURE SELECTION

Feature selection is the most common technique easy to use in the problem of text categorization, as mostly available for the feature selection process. However many unsupervised method are also used as a feature selection. [10]The terms selected after applying the wordnet, the metric weights all higher than the pre-specified threshold are selected as a key term. In our approach three different feature selection methods are used: tf-idf, tf-df, and tf2 to select representative terms of each document which are defined as follow:.
(1) **tf-idf** (term frequent - inverse document frequency): It is denoted as **tfi-dfij** and used for the measure of the importance of term **tj** within document **di**.

$$\mathbf{tf_i df_{ij}} = \frac{f_{ij}}{\sum_{j=1}^{m} f_{ij}} \mathbf{X} \log \left( \frac{|D|}{|\{d_i\, t_j \in d_i \in D\}|} \right)$$

Where $f_{ij}$ is the frequency of term $t_j$ in the document $d_i$, and the denominator is the total frequencies of all terms in document $d_i$ . |**D**| is the total number of documents in the document set **D**, and $|\{d_i\, t_j \in d_i \in D\}|$ is the number of documents containing term $t_j$

(2) **tf-df** (term frequency x document frequency): It is represented by $tfdf_{ij}$ and evaluated for the value calculated by dividing the term frequency (TF) by the document frequency (DF), where TF is the number of times a term $t_j$ appears in a document $d_i$ divided by the total frequencies of all terms in $d_i$ , and DF is used to determine the number of documents containing term $t_j$ divided by the total number of documents in the document set D.

$$tfdf_{ij} = \frac{TF}{DF} \text{ , where TF} = \frac{f_{ij}}{\sum_{j=1}^{m} f_{ij}} \text{ and}$$

$$\text{DF} = \frac{|\{d_i\, t_j \in d_i \in D\}|}{|D|}$$

(3) tf 2 : It is the multiplication of $tfidf_{ij}$ and $tfdf_{ij}$

and we denote it as $tf^2_{ij}$

$$tf^2_{ij} = tfidf_{ij} * tfdf_{ij}$$

TF-IDF, a vector space based representation is the common technique used for text processing. [4][5]In this representation, the term frequency for each word is normalized by the inverse document frequency, or IDF. It normalizes and reduces the weight of each any term which occurs frequently in the collection. This reduces the importance of common terms in the collection, ensuring that the matching of documents be more influenced by that of more discriminative words which have relatively low frequencies. After these weights of each term in each document have been calculated, those which have weights all higher than pre-specified thresholds are retained. Subsequently, these retained terms form a set of key terms for the document set D. This includes only meaningful key terms, which do not appear in a well-defined stop word list, and satisfy the Pre-defined minimum *tf-idf* threshold. threshold $\propto$, the minimum *tf-df* threshold $\beta$ and the minimum *tf2* threshold $\gamma$

Based on the above approach, the representation of a document can be derived by algorithm. The algorithm defined is as follow:

Algorithm 1. Document pre-processing algorithm
Input   1. A document set D= {d1, d2, d3….di,..dn}
        2. A well defines stop word list.
        3  The minimum tf-idf threshold $\propto$
        4. The minimum tf-df threshold $\beta$
        5. The minimum tf2 threshold $\gamma$
Output: The key terms of D, KD
Method: Step 1: Extract the term set TD= {t1, t2, t3…tj,..ts}
Step 2. Remove all stop words from TD.
Step 3. Apply word stemming for TD using porter stemmer algorithm
Step 4. WordNet Senses Disambiguation is applied as an English Database.
Step 5.  Find the Global unique words and obtain frequent key words from TD.
Step 6. Evaluate and retain all its weights tf-idf, tf-df, tf2
Step 7. Obtain the key term set KD

## IV. EXPERIMENTAL RESULTS

In this section, we experimentally evaluate the performance of the proposed algorithm. All the experiments have been performed on a Intel I5 Processor, Windows 7 OS machine with 8 GB memory.

### A. Datasets

On three to four different datasets the experiments are performed these datasets are widely adopted as standard benchmarks for the text categorization task. To find key terms, stop words were removed and stemming was performed. Documents then were represented as TF (Term Frequency) vectors, and unimportant terms were useless. This process implies a significant dimensionality reduction without loss of clustering performance. The statistics of these datasets, after the document pre-processing is summarized in table 1.

Table:1 Statistics of the datasets

| Data Sets | Number of Documents | Number of Natural classes | Class size | Average length of Documents |
|---|---|---|---|---|
| Reuters-21578 | 21578 | 04 | 986 | 12 |
| Classic 30 | 5697 | 03 | 2568 | 35 |
| 20 News Group (atheism) | 1000 | 02 | 1206 | 15 |
| 20 News Group (Computer Graphics) | 1000 | 03 | 1204 | 15 |
| 20 News Group ( Hardware) | 1000 | 02 | 945 | 10 |
| Reuters Transcription Subset (Wheat) | 20 | 02 | 185 | 19 |
| Reuters Transcription Subset (Trade) | 21 | 03 | 195 | 21 |
| Reuters Transcription Subset (Ship) | 20 | 02 | 185 | 19 |
| Reuters Transcription Subset (Money) | 20 | 02 | 185 | 19 |
| Reuters Transcription Subset (Grain) | 20 | 02 | 185 | 19 |
| Reuters Transcription Subset(Corn) | 20 | 02 | 185 | 19 |

They are heterogeneous in terms of document size, cluster size, number of classes, and document distribution. The smallest document set contains 20 documents, and the largest one contains 21578 documents. Table 1 summarizes the statistics of these datasets. The detailed information of these datasets is described as follows:

Reuters Transcription subset: This data was created by selecting 20 files each from the 10 largest classes
20 News Group: This dataset is a collection of 18,828 newsgroup documents, partitioned almost evenly across 20 different newsgroups. The 20 newsgroups collection has become a popular data set for experiments in text applications of machine learning techniques, such as text classification and text clustering. This dataset contain 6 classes with their subclasses.
Classic 30: This document set is a combination of the three classes CACM, CISI, and MED abstracts. Classic dataset includes 3203 CACM documents, 1460 CISI documents from information retrieval papers, and 1033 MEDLINE documents from medical journals [1].
Reuters 21578: For each text categorization, a specification of text is given of what categories that text belongs to. For the Reuters-21578 collection the documents are Reuter's newswire stories, and the categories are five different sets of content related categories. For each document, a human indexer decided which categories from which sets that document belonged to. Reuters-21578 text categorization test collection Distribution 1.0. Re0 includes 1504 documents with 13 classes [2].

### B. The Impact of Feature Selection

For Document Clustering, Feature selection is essential to make clustering task efficient and more accurate. The most important goal of feature selection is to extract topic related terms, which could present the content of each document. [11]For selecting the most representative features, the formulas are used to obtain the three weights and select these terms, which their weights are all higher than the pre- defined thresholds. The details of the tested dataset with their specified minimum threshold value are shown below in the table.

Table:2 - Datasets for minimum threshold $\propto = 0.028$ using WordNet

| Data Sets | Number of Terms | Number of Key terms Tf-idf | Percentage of term removed |
|---|---|---|---|
| Reuters -21578 | 462 | 447 | 3.24 |
| Classic 30 | 309807 | 257758 | 16.80 |
| 20 News Group (athesim) | 185684 | 42076 | 77.34 |
| 20 News Group (Computer Graphics) | 164667 | 50596 | 69.27 |
| 20 News Group (Hardware) | 138843 | 51851 | 62.65 |
| Reuters Transcription Subset (Wheat) | 1940 | 965 | 50.25 |
| Reuters Transcription Subset (Trade) | 3394 | 759 | 77.63 |
| Reuters Transcription Subset (Ship) | 1397 | 939 | 32.78 |
| Reuters Transcription Subset (Money) | 2755 | 706 | 74.37 |
| Reuters Transcription Subset (Grain) | 2102 | 1009 | 51.99 |
| Reuters Transcription Subset(Corn) | 2331 | 1084 | 53.34 |

Table:3 - Datasets for minimum threshold, $\beta = 0.01$ using WordNet

| Data Sets | Number of Terms | Number of Key terms Tf-Df | Percentage of term removed |
|---|---|---|---|
| Reuters -21578 | 462 | 70 | 84.84 |
| Classic 30 | 309807 | 8974 | 97.10 |
| 20 News Group (athesim) | 185684 | 3433 | 98.15 |
| 20 News Group (Computer Graphics) | 164667 | 5662 | 96.56 |
| 20 News Group (Hardware) | 138843 | 5008 | 96.39 |
| Reuters Transcription Subset (Wheat) | 1940 | 421 | 78.29 |
| Reuters Transcription Subset (Trade) | 3394 | 417 | 87.71 |
| Reuters Transcription Subset (Ship) | 1397 | 167 | 88.04 |
| Reuters Transcription Subset (Money) | 2755 | 272 | 90.12 |
| Reuters Transcription Subset (Grain) | 2102 | 392 | 81.35 |
| Reuters Transcription Subset(Corn) | 2331 | 555 | 76.19 |

Table: 4 - Datasets for minimum threshold, τ = 0.005 using WordNet

| Data Sets | Number of Terms | Number of Key terms Tf2 | Percentage of term removed |
|---|---|---|---|
| Reuters - 21578 | 462 | 55 | 88.10 |
| Classic 30 | 309807 | 8974 | 97.10 |
| 20 News Group (athesim) | 185684 | 1004 | 99.46 |
| 20 News Group (Computer Graphics) | 164667 | 1005 | 99.38 |
| 20 News Group (Hardware) | 138843 | 1000 | 99.27 |
| Reuters Transcription Subset (Wheat) | 1940 | 310 | 84.02 |
| Reuters Transcription Subset (Trade) | 3394 | 215 | 93.66 |
| Reuters Transcription Subset (Ship) | 1397 | 167 | 88.04 |
| Reuters Transcription Subset (Money) | 2755 | 166 | 93.97 |
| Reuters Transcription Subset (Grain) | 2102 | 290 | 86.20 |
| Reuters Transcription Subset (Corn) | 2331 | 437 | 81.25 |

## CONCLUSIONS

In this paper the key terms are extracted from the documents by using effective feature selection method tf-idf, tfdf, and tf2 based on its predefined threshold value and using background knowledge WordNet. In the total processes, we begin with the process of document pre-processing, removing the stop words, stemming by using porter stemmer algorithm, Wordnet lexical; database extracting global unique word, finding frequent word sets and further key terms are extracted from all documents by using feature selection methods. Based on the predefined threshold value and the key terms of tf-idf. tfdf, tf2 technique, the percentage of terms removed are evaluated. From the results it is observed that even if the percentage of terms removed is higher for tf2, tf-df than tf-idf but there is the higher possibility of data loss. The advantage of using tf-idf is that the dimensionality of the document gets reduced without data loss. Hence from the above process an effective feature selection method tf-idf is the better one for further mining process. Our future work focus on this tf-idf effective feature selection method using WordNet which can be used for attribute reduction by using rough set approach and k-means.